\documentclass[conference]{IEEEtran}
\IEEEoverridecommandlockouts
\usepackage{cite}
\usepackage{amsmath,amssymb,amsfonts}
\usepackage{algorithmic}
\usepackage{graphicx}
\usepackage{textcomp}
\usepackage{multirow}
\usepackage{subfigure}
\usepackage{xcolor}
\def\BibTeX{{\rm B\kern-.05em{\sc i\kern-.025em b}\kern-.08em
    T\kern-.1667em\lower.7ex\hbox{E}\kern-.125emX}}
\begin{document}

\title{Deep Transfer Tensor Factorization for\\
	Multi-View Learning
}

\author{\IEEEauthorblockN{Penghao Jiang}
\IEEEauthorblockA{\textit{The Australian National University} \\
Australia}
\and
\IEEEauthorblockN{Ke Xin}
\IEEEauthorblockA{\textit{The Australian National University} \\
	Australia}
\and
\IEEEauthorblockN{Chunxi Li}
\IEEEauthorblockA{\textit{The Australian National University} \\
	Australia}
}

\maketitle

\begin{abstract}
This paper studies the data sparsity problem in
multi-view learning. To solve data sparsity problem in multiview
ratings, we propose a generic architecture of deep transfer
tensor factorization (DTTF) by integrating deep learning and
cross-domain tensor factorization, where the side information
is embedded to provide effective compensation for the tensor
sparsity. Then we exhibit instantiation of our architecture by
combining stacked denoising autoencoder (SDAE) and CANDECOMP/
PARAFAC (CP) tensor factorization in both source and
target domains, where the side information of both users and
items is tightly coupled with the sparse multi-view ratings and
the latent factors are learned based on the joint optimization. We
tightly couple the multi-view ratings and the side information to
improve cross-domain tensor factorization based recommendations.
Experimental results on real-world datasets demonstrate
that our DTTF schemes outperform state-of-the-art methods on
multi-view rating predictions.
\end{abstract}

\begin{IEEEkeywords}
multi-view learning, tensor factorization, deep
learning, side information
\end{IEEEkeywords}

\section{Introduction}
With the data explosion in recent years, recommender
systems are becoming increasingly attractive. Traditional
single-view recommender systems typically operate on twodimensional
(2D) user-item ratings. In single-view recommender
systems, there are two primary categories of recommendation
algorithms: content-based methods and collaborative
filtering (CF) based methods, where matrix factorization
is effective in learning effective latent factors for users and
items \cite{1}. However, they cannot work well for multi-view recommender
systems that contain multiple view-specific ratings.

With the emergence of multi-modal or multi-aspect data,
multi-view recommendation becomes more and more important.
The list of applications ranges from social network
analysis to brain data analysis, and from web mining and
information retrieval to healthcare analytic \cite{2}, such as online
e-commerce websites and traveling portals. Figure 1 shows an
example in TripAdvisor, where customers can rate hotels by
using multiple view such as value, service, atmosphere, food
and overall, and meanwhile the information of customers and
hotels is also provided.

Prior multi-view techniques can be briefly classified into
three categories: heuristic neighborhood-based approaches
\cite{3}, aggregation-based approaches \cite{4}, and model-based approaches
\cite{5}. Heuristic neighborhood-based approaches attempt
to use various multi-view similarity metrics to collect
the neighbors of a targeted user, and then estimate unknown ratings based on the known ratings of those neighbors \cite{3},
\cite{6}. Aggregation-based approaches aim to build a mapping to
aggregate multiple view-specific ratings by assuming that there
is a certain relation between the overall rating and other viewspecific
ratings \cite{4}. Model-based approaches learn a predictive
model by leveraging the observed multi-criteria ratings and
then employing the model to execute prediction \cite{5}.

\begin{figure}[t]
\centering
\includegraphics[width=\columnwidth]{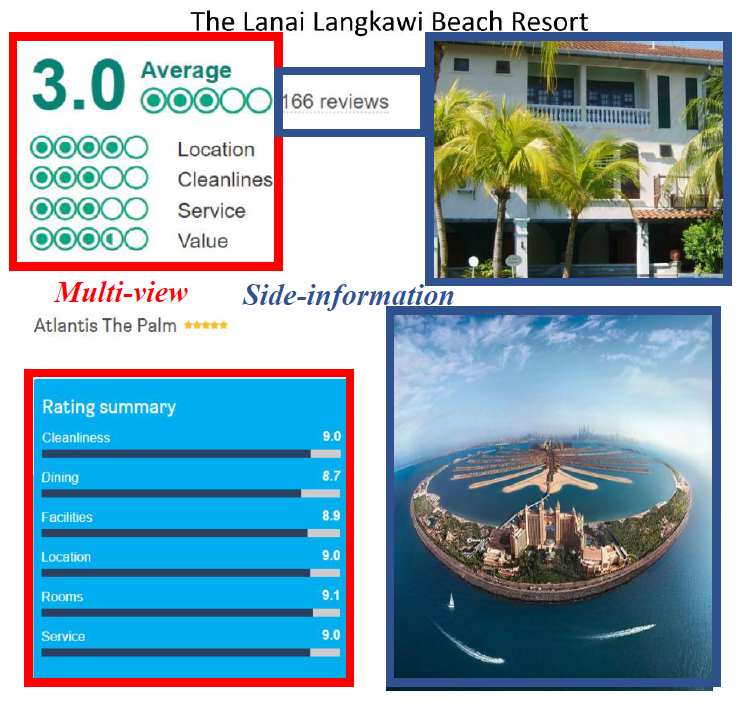}
\caption{Multi-view ratings and the side information from TripAdvisor}
\end{figure}

Tensor factorization is a milestone of model-based techniques
and many tensor factorization based techniques have
been developed for multi-view recommender systems \cite{7,8,9,10}.
Nevertheless, prior techniques suffer from the tensor sparsity problem. That being said, when the rating tensor is very sparse in real applications, the performance drops significantly. To overcome this problem in multi-view ratings, we attempt to 1) incorporate the \textit{side information} (or the auxiliary information) into the ratings to exploit prior features \cite{11,12,13,14} and 2) transfer or learn knowledge from relevant domains for crossdomain recommendations.

In this paper, we propose a generic architecture of deep transfer tensor factorization (DTTF) by integrating deep learning
and cross-domain tensor factorization, where the side
information is embedded to provide effective compensation for
the tensor sparsity. Then we exhibit instantiation of our architecture
by combining stacked denoising autoencoder (SDAE)
and CANDECOMP/PARAFAC (CP) tensor factorization in
both source and target domains, where the side information
of both users and items is tightly coupled with the sparse
multi-view ratings and the latent factors are learned based on
the joint optimization. The contribution of this paper can be
summarized as follows
\begin{itemize}
\item To solve data sparsity problem in multi-view ratings, we
propose a generic architecture to integrate deep structure
and cross-domain tensor factorization;
\item We present DTTF where cross-domain CP tensor factorization
is combined with four SDAEs in different domains
for users and items;
\item We tightly couple the multi-view ratings and the side
information to improve cross-domain tensor factorization
based recommendations.
\end{itemize}

\section{Related Work}
Multi-view learning (MTL) is an emerging direction in
machine learning which considers learning with multiple views
to improve the generalization performance. \cite{15} develop a
multi-view label embedding (MVLE) model by exploiting the
multi-view correlations. Nonlinear relationships usually exist
in real-world datasets, which have not been considered by most
existing methods. In order to address these challenges, a novel
model which simultaneously performs multi-view clustering
task and learns similarity relationships in kernel spaces is
proposed \cite{16}. In order to overcome the two limitations
\cite{17} propose a multi-task multi-view clustering algorithm in
heterogeneous situations based on Locally Linear Embedding
(LLE) and Laplacian Eigenmaps (LE) methods (L3EM2VC).
Comparing with existing methods that separately
cope with each view \cite{18} propose a supervised multi-view
feature learning framework to handle diverse views with a
unified perception. The proposed approach is compared to
different state-of-the-art Radiomics and multi-view solutions,
on different public multi-view datasets as well as on Radiomics
datasets \cite{19}. \cite{20} present a unified multi-view deep learning
framework to capture brain abnormalities associated with
seizures based on multi-channel scalp EEG signals. A novel
deep multi-view clustering model is proposed by uncovering
the hierarchical semantics of the input data in a layer-wise way
\cite{21}. The restricted Boltzmann machine (RBM) and extensions
are rarely used in the field of multi-view learning \cite{22}.

Matrix factorization has been widely used in single-view
recommender systems to solve the problem of personal information
overload \cite{1}. To mitigate the cold start and data
sparsity, it is inevitable for matrix factorization models to
exploit additional side information. Singh et al. \cite{23} have
integrated the side information into matrix factorization to
learn effective latent factors from sparse ratings, and shown an improved performance. Deep learning based matrix factorization
is designed to mitigate sparse ratings. A collaborative deep
learning based on Bayesian SDAE is proposed in \cite{13} that
attempts to incorporate the side information but only learns
latent representations for items. Deep collaborative filtering
is proposed based on marginalized denoising autoencoder to
learn latent representations for both items and users \cite{24}.
An alternative mode of incorporating the side information is
investigated in \cite{11}, which considers 2D ratings and the side
information in deep structure. Deep transfer structure shares
cross-domain information via hidden connections \cite{25} or learn
a common network via domain separation network. Different
from these scenarios, we study multi-view recommendations
via 3D tensor factorization instead.

Multi-view recommendation has been studied over decades
and can be briefly grouped into three categories: heuristic
neighborhood-based approaches \cite{26}, aggregation-based approaches
\cite{4}, and model-based approaches \cite{5}. Tensor factorization
is a milestone of model-based techniques.

Heuristic neighborhood-based approaches attempt to use
various multi-view similarity metrics to collect the user neighbors
and then predict based on known ratings of those neighbors.
Different techniques are proposed to find the best neighbors,
including a multi-dimensional distance metric \cite{3}, a preference
lattice based on user view preferences \cite{27}, and multiview
Euclidean distance \cite{6}. Aggregation-based approaches
build a mapping to aggregate multiple view-specific ratings for
prediction by assuming that there is a certain relation between
overall ratings and individual ratings. Lakiotaki et al. \cite{4} proposes
a utility additive method to aggregate the marginal user'
preferences on the given criteria. Jannach et al. \cite{28} uses a
support vector regression to learn relative importance of viewspecific
ratings and then combines regression models for users
and items to predict unknown ratings. A view chain-based
method is presented in \cite{29} to aggregate the multi-dimensional
ratings for recommendations by considering the dependence
among multiple view ratings. The empirical results of the
comparative analysis of their performance are presented \cite{30}.
Since outputs of expert systems directly dependent on input
signals; interventions to the inputs coherently cause failures
on productions of such systems. \cite{31} examine shilling attack
strategies against multi-criteria preference collections, how to
extend well-known attack scenarios against these systems,
and propose an alternative attacking scheme. \cite{32} introduce a
tensor factorization method to handle three-dimensional useritem-
criterion rating data. \cite{33} propose a utility-based multicriteria
recommendation algorithm, in which \cite{33} learn the
user expectations by different learning-to-rank methods. The
resulting compressed vectors constitute latent multi-criteria
ratings that \cite{34} use for the recommendation purposes via
standard multi-criteria recommendation methods. \cite{35} propose
a novel multi-criteria collaborative filtering model based on
deep learning. The personalized recommendation technology
can establish user files through the user's behavior and other
information, and automatically recommend the items that
best match the user's preferences, thus effectively reducing the information overload problem. Based on this \cite{36} study
the personalized recommendation algorithm based on user
preferences in mobile e-commerce.

Model-based approaches aim to learn a predictive model
and then employ the model to estimate the ratings. Many
techniques have been proposed for recommendations, including
a probabilistic mixture algorithm \cite{5}, an adaptive neurofuzzy
inference and self-organizing map clustering \cite{37}, and a
multi-linear singular value decomposition \cite{38}. Tensor factorization
is a milestone of model-based approaches and various
methods have been developed for wide applications \cite{2}. A
tensor factorization based ranking is presented in \cite{39} to
predict personalized tags for users. A high-order singular value
decomposition (HOSVD) method is used in \cite{7} to deal with
contextual information for context-aware recommendations,
where the limitation is that it primarily works for categorical
context variables. Rendle et al. \cite{40} proposes a factorization
machine method by extending HOSVD. And Zhang et al.
\cite{41} presents tensor singular value decomposition (t-SVD) that
can perfectly recover a tensor with low tubal-rank under the
certain tensor standard incoherent condition. Based on classic
matrix factorization, Bhargava et al. \cite{9} tackles context-aware
collaborative recommendation by tensor while Yao et al. \cite{10}
presents an application in point-of-interest recommendations.
Chen et al. \cite{42} proposes deep tensor factorization to integrate
deep representation learning and tensor factorization for multiview
recommendations.

In this paper, we integrate tensor factorization and deep
structure to incorporate the side information, and link tensor
factorization in source domain with that in the target domain.

\section{Preliminary and Overview}
This paper aim to cope with the multi-view recommendation problem, similar to some existing works \cite{3,43}. Generally, multi-view recommender systems refer to the systems that leverage multiple categories of ratings based on various specific view in addition to the overall \textit{user-item} ratings to implement recommendation tasks. Figure 2 shows an example of a 3D \textit{user-item-view} rating tensor $\mathbf{R}$, where each user rates on various view of a given item, and the mark ``?" means unobserved ratings. And the rating tensor is extremely sparse with $I$ users, $J$ items, $L$ view in this paper. Each rating $r_{i j l}$ in the tensor $\mathbf{R}$ corresponds to user $i$ rates on the view $l$ of item $j$. Given the sparse third-order user-item-view rating tensor $\mathbf{R}$, the side information matrix $\mathbf{M}$ for users and $\mathbf{N}$ for items, the goal is to learn user latent factors $\mathbf{U}$, item latent factors $\mathbf{V}$ and view latent factors $\mathbf{C}$, and then predict the unobserved ratings in $\mathbf{R}$.

\begin{figure}[t]
	\centering
	\includegraphics[width=1.05\columnwidth]{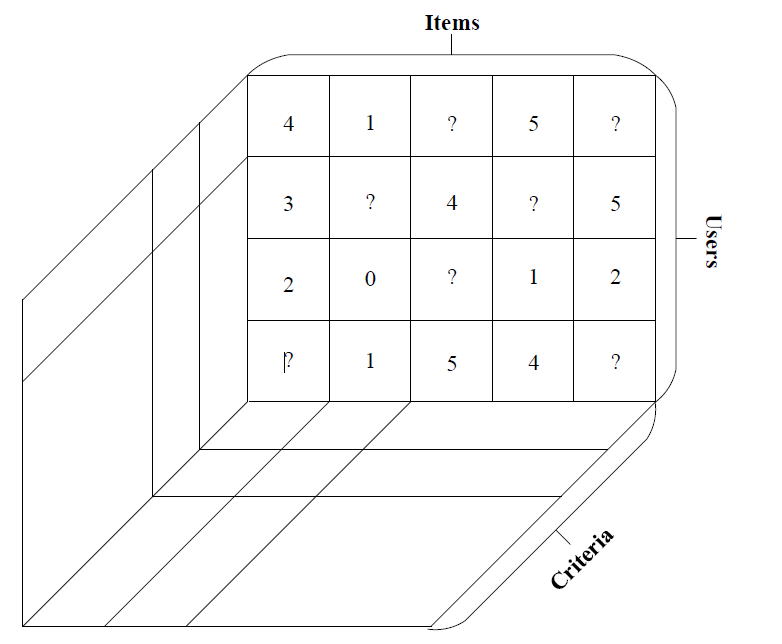}
	\caption{Rating tensor}
\end{figure}
\vspace{3mm}
\begin{figure}[t]
	\centering
	\includegraphics[width=1\columnwidth]{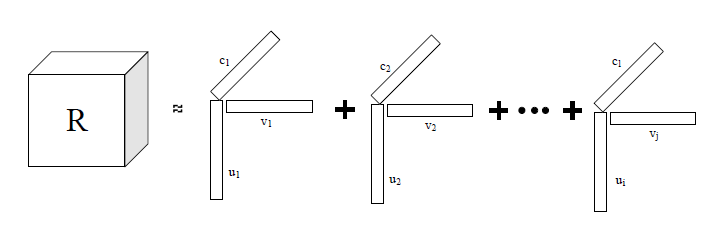}
	\caption{The CP tensor factorization}
\end{figure}
\begin{figure*}[t]
	\centering
	\includegraphics[width=\textwidth]{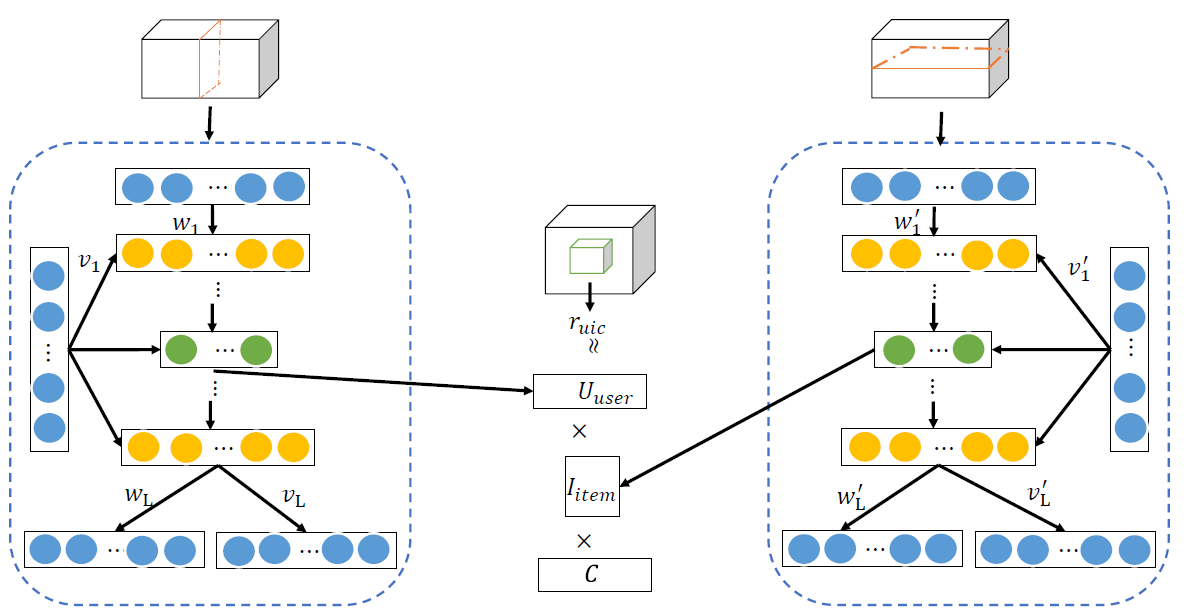}
	\caption{The structure of the proposed DTTF}
\end{figure*}

Given 3D \textit{user-item-view} ratings, tensor factorization is to
map users, items, and view into a joint latent factor space
so that the users' preferences on specific view of items can
be formulated as the inner products of corresponding latent
factor vectors in the space. The CP is widely used as a
tensor factorization paradigm due to its key advantage of linear
complexity. We adopt the CP to decompose the rating tensor
in this work.

Figure 3 shows the CP tensor factorization, where a \textit{user-item-view} rating tensor $\mathbf{R} \in \mathbb{R}^{I * J * L}$ can be decomposed into a sum of rank-one tensors across the whole set of users. So, we have
\begin{equation}
\arg \min _{\mathbf{U}, \mathbf{V}, \mathbf{C}} I\|\mathbf{R}-\mathbf{U} \otimes \mathbf{V} \otimes \mathbf{C}\|^2
\end{equation}
where $\mathbf{U} \in \mathbb{R}^{I \times K}, \mathbf{V} \in \mathbb{R}^{J \times K}$ and $\mathbf{C} \in \mathbb{R}^{L \times K}$ represent the latent factor matrix for users, items and view, respectively; $K$ is the dimension of latent factor space; and the operator $\otimes $ detnotes the outer product of latent factor vectors in the corresponding matrix.

In this paper, we attempt to propose a novel architecture
to combine deep structure and cross-domain CP tensor factorization,
where deep structure deals with either only the side
information or both the ratings and the side information, and
tensor factorization deals with the 3D \textit{user-item-view} ratings.

Four deep structures are designed for users and items in
both source and target domains, where the side information of
users (or items) is involved as an input, and the transformation
of ratings are either taken as one more input or not included
(dashed line). The effective latent representation is learned by
jointly optimizing deep network and latent factors from tensor
factorization. The tightly coupled side information provides a compensation for tensor factorization, so the proposed DTTF
could mitigate the tensor sparsity problem.

For convenient description, define by $\mathcal{D}_s$ the source domain and $\mathcal{D}_t$ the target domain. And the domain indices are denoted as $d \in\{s, t\}$. In a recommendation setting, the user-item-view matrix $\mathbf{R}_d \in \mathbb{R}^{I_d \times J_d \times L}$ can be decomposed as a sum of rank1 tensors across all users. So, we have
\begin{equation}
\arg \min _{\mathbf{U}_d, \mathbf{V}_d, \mathbf{C}}\left\|\mathbf{R}_d-\mathbf{U}_d \otimes \mathbf{V}_d \otimes \mathbf{C}\right\|^2,
\end{equation}
where $\mathbf{U}_d \in \mathbb{R}^{I_d \times K}, \mathbf{V}_d \in \mathbb{R}^{J_d \times K}$ and $\mathbf{C} \in \mathbb{R}^{L \times K}$ represent the latent factor matrix for users, items and view, respectively; $K$ is the dimension of latent factor space; and $\otimes$ denotes the outer product of latent factor vectors in the corresponding matrix. The source domain $\mathcal{D}_s$ is connected with the target domain $\mathcal{D}_t$ via common latent factors $\mathbf{C}$.

\section{Deep Transfer Tensor Factorization}
In this section, DTTF instantiation is presented in detail
based on the generic architecture.

\subsection{DTTF Scheme}
The specific DTTF scheme is composed of several components:
a SDAE for users, a SDAE for items and tensor
factorization in both domains, as shown in Figure 4. In
DTTF, the SDAE only takes the side information as the sole
input, similar to \cite{44}, where the multi-view ratings are not considered.

Considering the SDAE for users in Figure 4 , the representation $h_{d, l}^{(u)}$ at each hidden layer and the output at layer $L^{(u)}$ can be obtained as
\begin{equation}
\begin{aligned}
& \mathbf{h}_{d, l}^{(u)}=g\left(\mathbf{W}_{d, l}^{(u)} \mathbf{h}_{d, l-1}^{(u)}+\mathbf{b}_{d, l}^{(u)}\right) \\
& \hat{\mathbf{p}}_{d, i}^{(u)}=f\left(\mathbf{W}_{d, L^{(u)}}^{(u)} \mathbf{h}_{d, L^{(u)}}^{(u)}+\mathbf{b}_{d, L^{(u)}}^{(u)}\right),
\end{aligned}
\end{equation}
where $l \in\{1,2, \cdots, L_d^{(u)}-1\} ; g(\cdot)$ and $f(\cdot)$ are activation functions for the hidden and output layers. The corrupted side information $\tilde{\mathbf{p}}_{d, i}^{(u)}$ is the input to the first layer, $\mathbf{h}_{d, r}^{\left(u_i\right)}$ denotes deep representations from the middle layer and $\hat{\mathbf{p}}_{d, i}^{(u)}$ denotes the output of the users' SDAE. Similar results can be obtained for the items' SDAE by replacing $(u)$ with $(v)$.

As observed in Figure 4 , the users' SDAE takes as input the side information of users to learn the latent representation $\mathbf{h}_{d, r}^{\left(u_i\right)}$ that is used to compensate latent factor vectors $\mathbf{u}_{d, i}$ in tensor factorization. And the items' SDAE takes as input the side information of items to learn latent representation $\mathbf{h}_{d, r}^{\left(v_j\right)}$ that is used to compensate the latent factor vectors $\mathbf{v}_{d, j}$ in tensor factorization.

\subsubsection{Loss Function}
DTTF learns users' latent factors, items' latent factors and view latent factors through the following objective function
\begin{equation}
\min _{\Theta} \mathcal{J}=\mathcal{L}_t+\mathcal{L}_r+\mathcal{L}_a+\lambda f_{ {reg }}
\end{equation}
where the overall loss function $\mathcal{J}$ consists of four components: the loss of tensor factorization $\mathcal{L}_t$, the reconstruction cost of the side information $\mathcal{L}_r$, the approximation error between deep representation and latent factors $\mathcal{L}_a$, and the regularization term $f_{ {reg }}$ that prevent overfitting.

The first term $\mathcal{L}_t$ denotes the loss of factorization on a sparse rating tensor
\begin{equation}
\min _{\boldsymbol{\theta}_t} \mathcal{L}_t=\sum_{d \in(s, t)}\left\|\mathbf{I}_{\mathbf{d}} \odot\left(\mathbf{R}_{\mathbf{d}}-\mathbf{U}_{\mathbf{d}} \otimes \mathbf{V}_{\mathbf{d}} \otimes \mathbf{C}\right)\right\|^2,
\end{equation}
where $\theta_t=\left\{\mathbf{U}_{\mathbf{d}}, \mathbf{V}_{\mathbf{d}}, \mathbf{C}\right\}$; the binary tensor $\mathbf{I}_d$ is an indicator of sparsity, in which each element indicates whether the corresponding rating is observed $(=1)$ or not $(=0)$; $\otimes$ means the outer product of latent factor vectors in the corresponding matrix; and $\odot$ is the element-wise production.

Secondly, the reconstruction cost of the side information for
both users and items can be expressed as
\begin{equation}
\small\min _{\theta_r} \mathcal{L}_r=\sum_d\left[\alpha_d \sum_i\left(\mathbf{p}_{d, i}^{(u)}-\hat{\mathbf{p}}_{d, i}^{(u)}\right)^2+\beta_d \sum_j\left(\mathbf{p}_{d, j}^{(v)}-\hat{\mathbf{p}}_{d, j}^{(v)}\right)^2\right],
\end{equation}
where $\theta_r=\left\{\mathbf{W}_d^u, \mathbf{b}_d^u, \mathbf{W}_d^v, \mathbf{b}_d^v\right\}, \alpha_d$ and $\beta_d$ are penalty parameters.

Furthermore, the approximation error between deep representation and latent factor vectors for both users and items can be expressed as
\begin{equation}
\small
\min _{\theta_a} \mathcal{L}_a=\sum_d\left[\rho_d \sum_i\left(\mathbf{u}_{d, i}-\mathbf{h}_{d, r}^{\left(u_i\right)}\right)^2+\gamma_d \sum_j\left(\mathbf{v}_{d, j}-\mathbf{h}_{d, r}^{\left(v_j\right)}\right)^2\right],
\end{equation}
where $\boldsymbol{\theta}_a=\left\{\mathbf{U}_d, \mathbf{V}_d, \mathbf{W}_d^{(u)}, \mathbf{b}_d^{(u)}, \mathbf{W}_d^{(v)}, \mathbf{b}_d^{(v)}\right\}, \rho_d$ and $\gamma_d$ are penalty parameters.

The last term denotes the regularization term $f_{ {reg }}$ as
\begin{equation}
\begin{aligned}
f_{ {reg }} & =\sum_d\left(\sum_i\left\|\mathbf{u}_{d, i}\right\|^2+\sum_j\left\|\mathbf{v}_{d, j}\right\|^2\right) \\
& +\sum_d\left(\left\|\mathbf{W}_d^{(u)}\right\|^2+\left\|\mathbf{W}_d^{(v)}\right\|^2+\left\|\mathbf{b}_d^{(u)}\right\|^2+\left\|\mathbf{b}_d^{(v)}\right\|^2\right),
\end{aligned}
\end{equation}
and the overall $\Theta=\theta_t \cup \theta_r \cup \theta_a$ in (4).

\subsubsection{Optimization}
To solve this problem, the alternative optimization algorithm is considered by utilizing the following three-step procedure.\\
\textit{Step I}: Given all weights $\mathbf{W}_d$ and biases $\mathbf{b}_d$, the gradients of $\mathcal{J}$ in (4) with respect to $\mathbf{u}_{d, i}, \mathbf{v}_{d, j}$, can be obtained as
\begin{equation}
\begin{aligned}
\frac{\partial \mathcal{J}}{\partial \mathbf{u}_{d, i}} & =-\sum_j \sum_l \mathcal{I}_{d, i j l}\left(r_{d, i j l}-\mathbf{u}_{d, i} \mathbf{v}_{d, j} \mathbf{c}_l\right)\left(\mathbf{v}_{d, j} \mathbf{c}_l\right) \\
& +\rho_d\left(\mathbf{u}_{d, i}-\mathbf{h}_{d, r}^{\left(u_i\right)}\right)+\lambda \mathbf{u}_{d, i} \\
\frac{\partial \mathcal{J}}{\partial \mathbf{v}_{d, j}} & =-\sum_i \sum_l \mathcal{I}_{d, i j l}\left(r_{d, i j l}-\mathbf{u}_{d, i} \mathbf{v}_{d, j} \mathbf{c}_l\right)\left(\mathbf{u}_{d, i} \mathbf{c}_l\right) \\
& +\gamma_d\left(\mathbf{v}_{d, j}-\mathbf{h}_{d, r}^{\left(v_j\right)}\right)+\lambda \mathbf{v}_{d, j}
\end{aligned}
\end{equation}
\textit{Step II}: Fixed the users' latent factors $\mathbf{U}_d$ and the items' latent factors $\mathbf{V}_d, d \in\{s, t\}$, the common latent factors $\mathbf{C}\left(\mathbf{c}_l\right)$ can be updated by
\begin{equation}
\begin{aligned}
\frac{\partial \mathcal{J}}{\partial \mathbf{c}_l} & =-\sum_d \sum_i \sum_j \mathcal{I}_{d, i j l}\left(r_{d, i j l}-\mathbf{u}_{d, i} \mathbf{v}_{d, j} \mathbf{c}_l\right)\left(\mathbf{u}_{d, i} \mathbf{v}_{d, j}\right) \\
& +\lambda \mathbf{c}_l,
\end{aligned}
\end{equation}
where the binary $\mathcal{I}_{d, i j l}$ indicates whether the corresponding rating is observed $(=1)$ or not $(=0)$.\\
\textit{Step III}: Fixed the latent factors $\mathbf{U}, \mathbf{V}$ and $\mathbf{C}$, all weights $\mathbf{W}$ and biases $\mathbf{b}$ of both SDAEs can be learned by backpropagation with stochastic gradient decent (SGD) method
\begin{equation}
\begin{aligned}
\frac{\partial \mathcal{J}}{\partial \mathbf{W}_d^{(u)}} & =-\rho_d \sum_i\left(\mathbf{u}_{d, i}-\mathbf{h}_{d, r}^{\left(u_i\right)}\right) \frac{\partial \mathbf{h}_{d, r}^{\left(u_i\right)}}{\partial \mathbf{W}_d^{(u)}} \\
& +\alpha_d \sum_i\left(\mathbf{p}_{d, i}^{(u)}-\hat{\mathbf{p}}_{d, i}^{(u)}\right) \frac{\partial \hat{\mathbf{p}}_{d, i}^{(u)}}{\partial \mathbf{W}_d^{(u)}}+\lambda \mathbf{W}_d^{(u)} \\
\frac{\partial \mathcal{J}}{\partial \mathbf{W}_d^{(v)}} & =-\gamma_d \sum_j\left(\mathbf{v}_{d, j}-\mathbf{h}_{d, r}^{\left(v_j\right)}\right) \frac{\partial \mathbf{h}_{d, r}^{\left(v_j\right)}}{\partial \mathbf{W}_d^{(v)}} \\
& +\beta_d \sum_j\left(\mathbf{p}_{d, j}^{(v)}-\hat{\mathbf{p}}_{d, j}^{(v)}\right) \frac{\partial \hat{\mathbf{p}}_{d, j}^{(v)}}{\partial \mathbf{W}_d^{(v)}}+\lambda \mathbf{W}_d^{(v)}
\end{aligned}
\end{equation}
and $\frac{\partial \mathcal{J}}{\partial \mathbf{b}_d^{(u)}}$ and $\frac{\partial \mathcal{J}}{\partial \mathbf{b}_d^{(v)}}$ can be easily obtained by replacing $\mathbf{W}_d$ with $\mathbf{b}_d$ in (11). Iterate three steps above until convergence.

\begin{table*}[t]
\centering
\caption{Performance comparison of various methods in terms of RMSE.}
\begin{tabular}{|c|c|c|c|c|c|c|c|c|c|c|c|c|}
	\hline
	\multirow{2}{*}{Algorithm}&\multicolumn{3}{c|}{TA12M (s) vs TA20M (t)}&\multicolumn{3}{c|}{TA20M (s) vs TA12M (t)}&\multicolumn{3}{c|}{TA30M (s) vs TA100M (t)}&\multicolumn{3}{c|}{TA100M (s) vs TA30M (t)}\\\cline{2-13}
	&60\%&80\%&95\%&60\%&80\%&95\%&60\%&80\%&95\%&60\%&80\%&95\%\\\hline\noalign{\vspace{1mm}}
	\hline AFBM & 1.219 & 1.167 & 1.096 & 1.178 & 1.053 & 1.045 & 0.787 & 0.784 & 0.716 & 0.950 & 0.937 & 0.934 \\
	\hline CMF & 1.184 & 1.140 & 1.130 & 1.274 & 1.058 & 1.038 & 0.713 & 0.693 & 0.653 & 0.855 & 0.832 & 0.810 \\
	\hline DCF & 1.164 & 1.094 & 1.036 & 1.163 & 1.069 & 1.031 & 0.668 & 0.643 & 0.628 & 0.794 & 0.772 & 0.743 \\
	\hline HCF & 1.128 & 1.073 & 1.030 & 1.089 & 1.066 & 1.016 & 0.653 & 0.639 & 0.617 & 0.761 & 0.734 & 0.727 \\
	\hline t-SVD & 1.181 & 1.075 & 1.039 & 1.151 & 1.040 & 0.961 & 0.620 & 0.598 & 0.579 & 0.671 & 0.660 & 0.644 \\
	\hline DTF & 1.082 & 1.049 & 1.029 & 1.022 & 1.016 & 0.869 & 0.610 & 0.597 & 0.578 & 0.668 & 0.642 & 0.632 \\
	\hline DTTF & \textbf{1.037}&\textbf{0.963}&\textbf{0.868}&\textbf{0.930}&\textbf{0.899}&\textbf{0.851}&\textbf{0.604}&\textbf{0.587}&\textbf{0.567}&\textbf{0.660}&\textbf{0.628}&\textbf{0.619} \\
	\hline
\end{tabular}
\end{table*}

\begin{figure*}[t]
\centering
\subfigure[TA-12M]{
\begin{minipage}{0.8\columnwidth}
\includegraphics[width=\columnwidth]{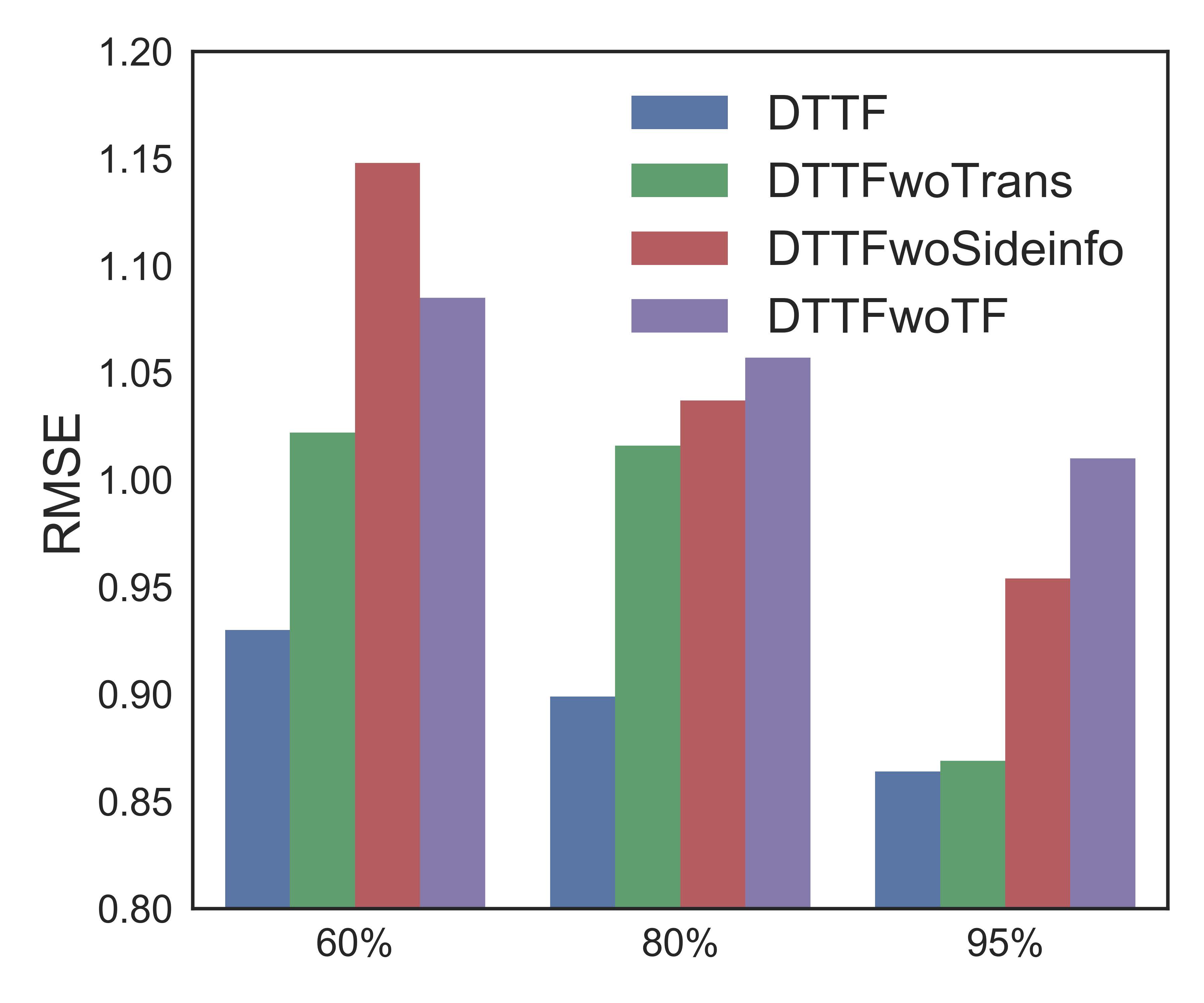}
\end{minipage}}
\subfigure[TA-20M]{
	\begin{minipage}{0.8\columnwidth}
		\includegraphics[width=\columnwidth]{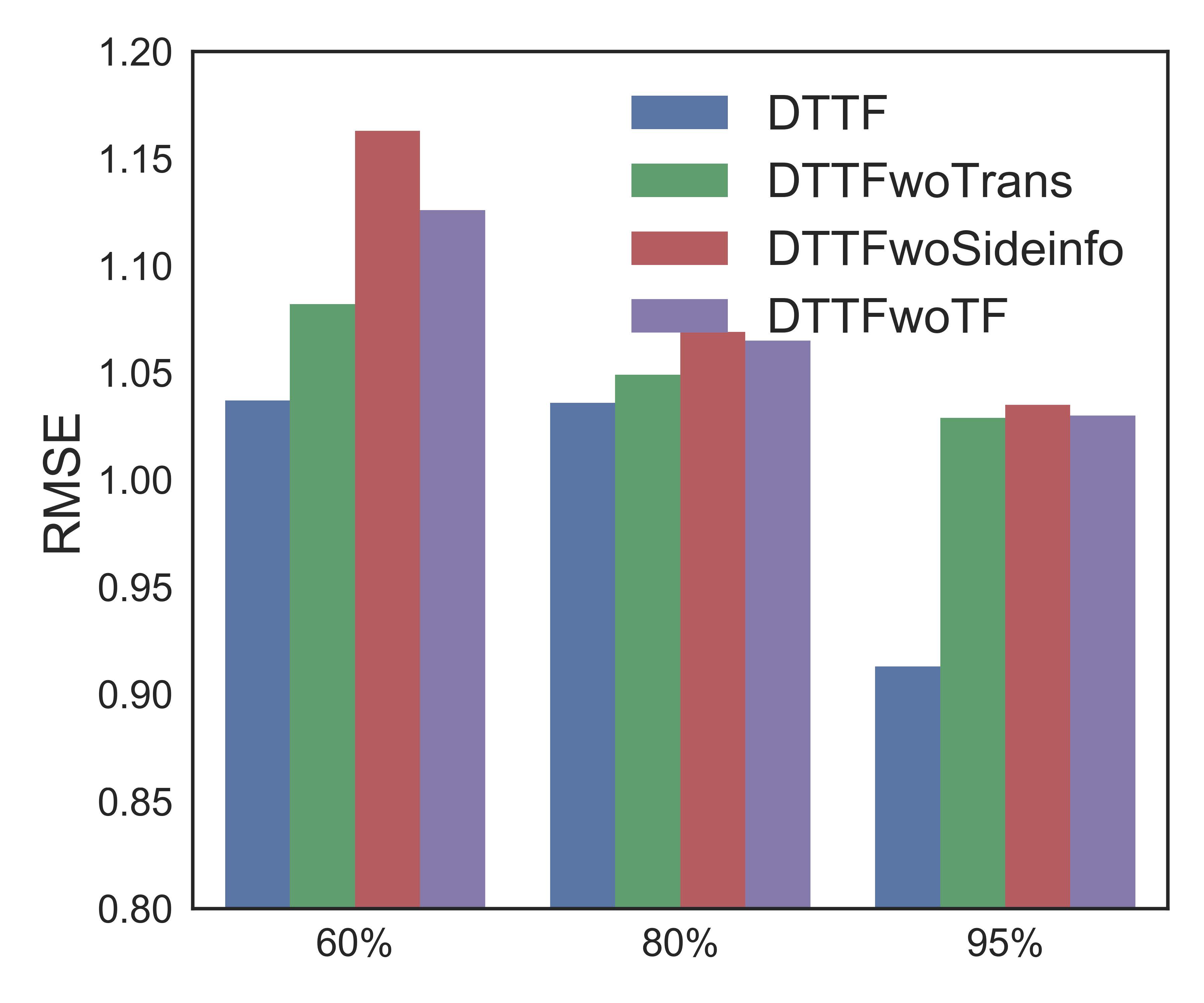}
\end{minipage}}
\subfigure[TA-30M]{
	\begin{minipage}{0.8\columnwidth}
		\includegraphics[width=\columnwidth]{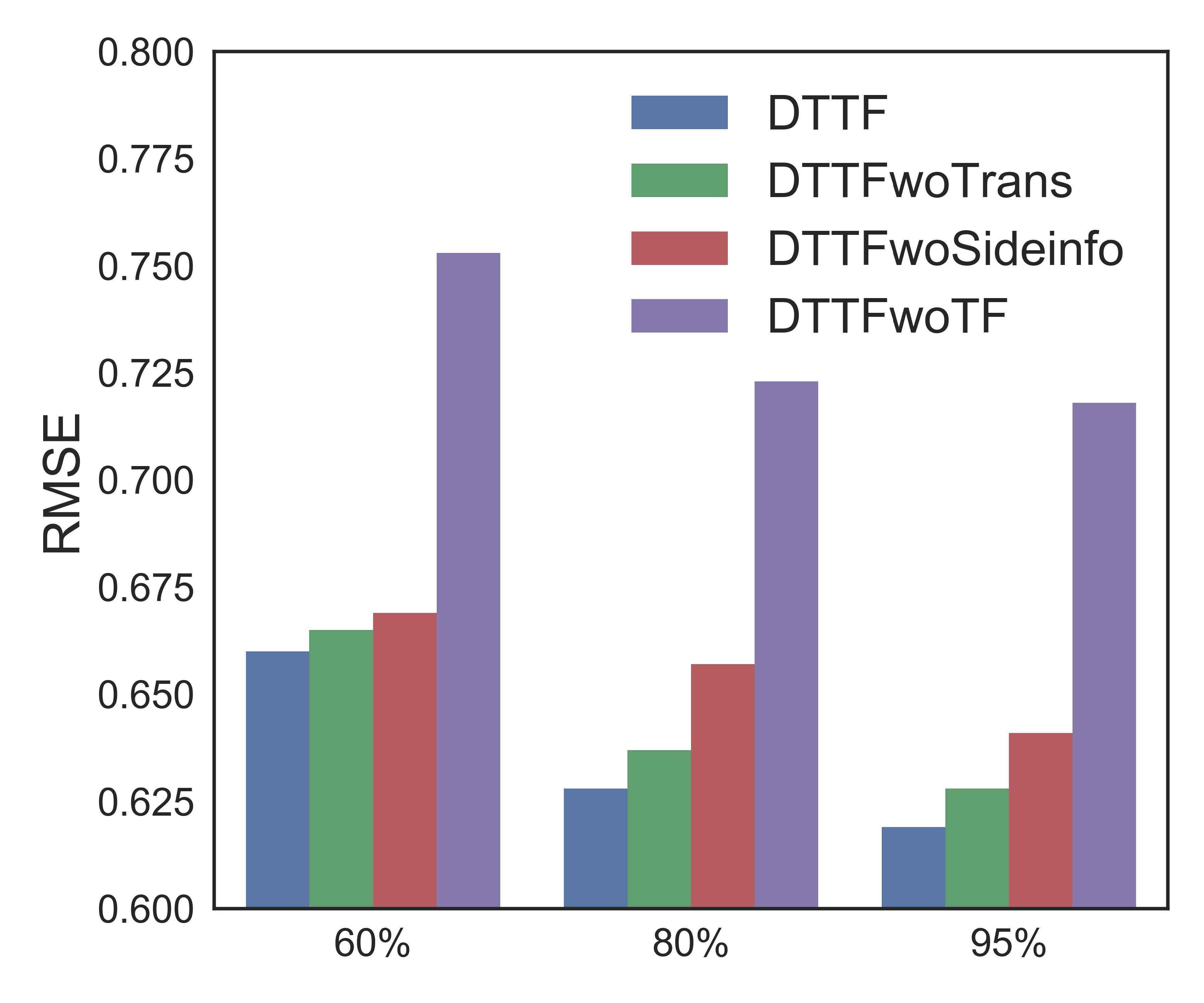}
\end{minipage}}
\subfigure[TA-100M]{
	\begin{minipage}{0.8\columnwidth}
		\includegraphics[width=\columnwidth]{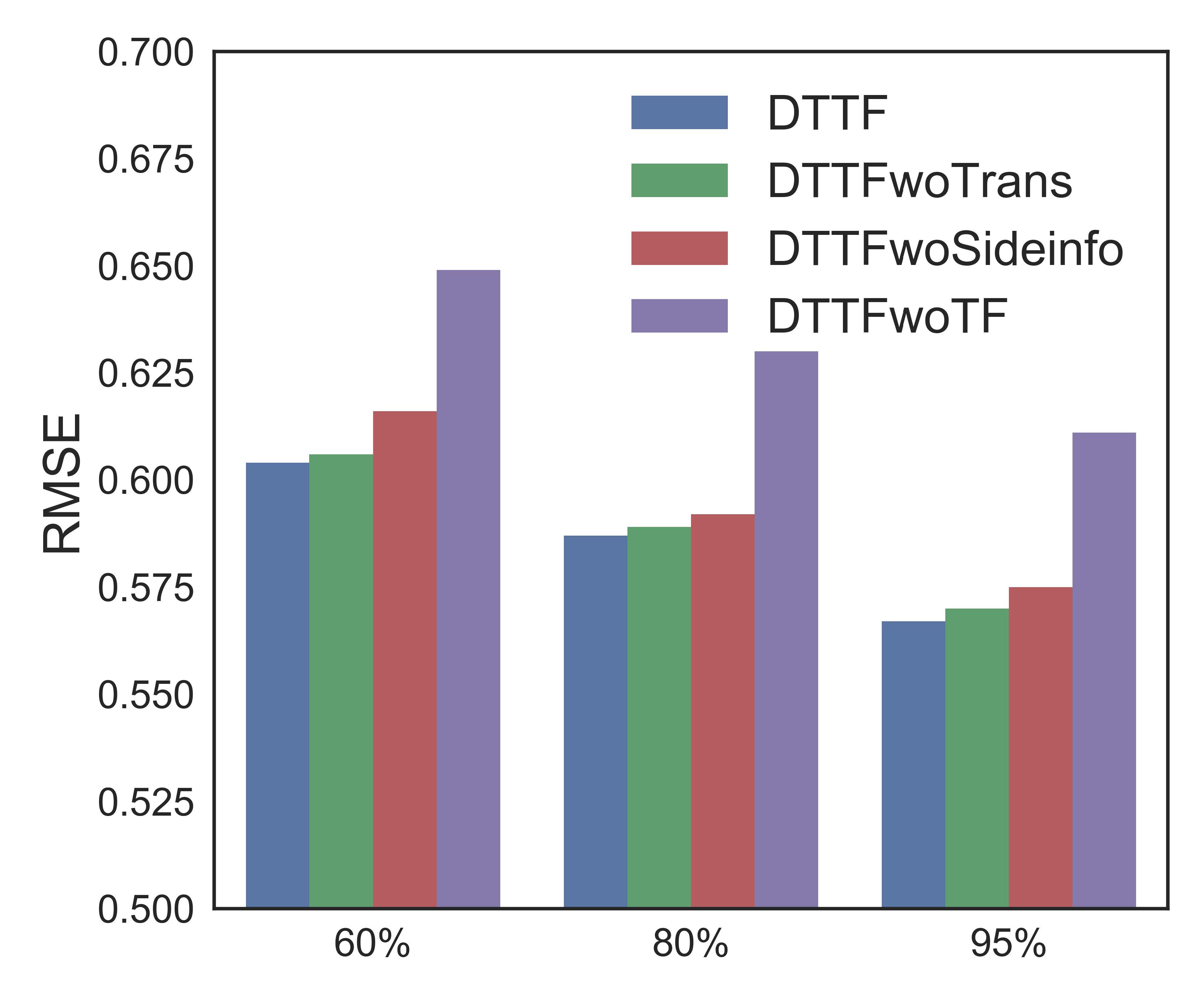}
\end{minipage}}
\caption{Ablation Test Results of DTTF on four datasets.}
\end{figure*}

\section{Experiments}
\subsection{Experiment Setup}
To evaluate various algorithms, we use four public datasets,
two from TripAdvisor (TA) and two from RateBeer (RB).
All these datasets are commonly used for evaluating the
performance of recommender systems \cite{28,45}. They are
different datasets without any overlap and independent of each other.

\begin{itemize}
\item \textit{TripAdvisor-12M (TA-12M)}: This dataset contains
181,411 records given by 1,750 users based on 4 view
including \textit{value}, \textit{location}, \textit{service}, and \textit{overall} for 3,546 hotels. Each user gave at least 2 ratings. The sparsity level of the dataset is around 99.26\%.
\item \textit{TripAdvisor-20M (TA-20M)}: This dataset contains 63,945
records given by 2,246 users based on 4 view including
\textit{value}, \textit{location}, \textit{service}, and \textit{overall} for 3,033 hotels.
The sparsity level of the dataset is around 99.76\%.
\item \textit{RateBeer-30M (RB-30M)}: This dataset contains
1,326,451 records given by 2,167 users for 3,109
beers based on 5 view including \textit{appearance}, \textit{aroma},
\textit{palate}, \textit{taste} and \textit{overall}. The sparsity level of the
dataset is around 96.20\%.
\item \textit{RateBeer-100M (RB-100M)}: This dataset contains 2,294,766 records given by 1,771 users for 2,627 beers based on 5 view including \textit{appearance}, \textit{aroma}, \textit{palate},
\textit{taste} and \textit{overall}. The sparsity level of the dataset is
around 90.13\%.
\end{itemize}

For TA datasets, the user and item additional matrices are
generated similarly as RB datasets. And the length of the
resulting binary vector is 106 for users and 134 for items.

In our experiments, five-fold cross validation was applied to
each dataset, and we use the root mean squared error (RMSE),
the Hit Ratio (HR) and Normalized Discounted Cumulative
Gain (NDCG) \cite{46} as the evaluation metric.

\subsection{Baseline}
In order to evaluate the performance, we consider the
following baselines in our experiments:
\begin{itemize}
\item \textit{AFBM}: Aggregation function based method \cite{26} employs a matrix factorization to deal with the observed user-view ratings.
\item \textit{CMF}: Collective matrix factorization \cite{23} is a model which simultaneously decomposes the ratings and the side information.HCF: HCF is a hybrid collaborative filtering model \cite{11}
which unifies aSDAE model with matrix factorization.
\item \textit{DCF}: Deep collaborative filtering \cite{44} is a recommendation
model which combines probabilistic matrix factorization
with marginalized denoising stacked autoencoders to
achieve recommendation.
\item \textit{t-SVD}: Tensor Singular Value Decomposition \cite{41} is a
model to generalize MF approaches to higher dimensional
multi-view recommendations.
\item \textit{DTF}: Deep tensor factorization \cite{42} is a model to integrate
deep representation learning and tensor factorization
for multi-view recommendations.
\end{itemize}

\subsection{Comparison Experimental Results}
We evaluate our proposed DTTF on four datasets in comparison
to state-of-the-art recommendation baselines.

Table I illustrates the performance of all methods in terms of
the average RMSE, where the lowest RMSE in each dataset
is highlighted in boldface and the second lowest RMSE is
highlighted in italic boldface. The proposed DTTF clearly
outperform all baselines in terms of RMSE, in which DTTF
achieves the \textit{best} performance for all cases.

Specifically, it is observed that HCF, DCF and CMF outperform
AFBM in general cases, and DTTF schemes outperform
t-SVD, which demonstrates the effectiveness of incorporating
the side information in either 2D rating matrix or 3D rating
tensor. That DTTF, HCF and DCF outperform CMF indicates
that deep structure can acquire better features of the side
information. HCF, DCF, CMF and AFBM only consider the
correlation between arbitrary two of three dimensions so
DTTF and t-SVD outperform these methods. That DTTF, DTF
outperform DCF and HCF indicates that tensor factorization
methods effectively learn the intrinsic interactions among three
dimensions, which are a good fit for multi-view recommender
systems. And DTTF outperform DTF which only consider
single-domain dataset, validating the effectiveness of crossdomain
learning in multi-view recommendations.

\subsection{Ablation Analysis}
The comparison results in terms of per evaluation metrics
indicate that the proposed DTTF clearly outperform the wellestablished
baselines.

To justify the efficiency of our architecture design, a careful
ablation study is conducted. Specifically, we remove the
knowledge transfer from either DTTF and name it as DTTFwoTrans;
we remove the deep structure of side information
from DTTF and name it as DTTFwoSideinfo; we remove the
tensor factorization component from DTTF and name it as
DTTFwoTF.

The test results in terms of RMSE are shown in Figure 5 and
a few observations are worth being highlighted as follows: 1)
The best performance on each dataset is obtained by the complete
DTTF, indicating that each of components contributes
to the effectiveness and robustness of the whole model; 2)
The RMSE of DTTFwoSideinfo in TA datasets is significantly
higher than others, indicating that the incorporation of side information is crucial for the sparsity problem in multi-view recommender systems. 3) The RMSE of DTTFwoTF in RB datasets is significantly higher than others. One reason is that the density of the TA-12M dataset $(0.73 \%)$ and TA20M dataset $(0.23 \%)$ is much lower than RB-30M $(3.80 \%)$ and RB-100M dataset $(9.87 \%)$. Another possible reason is that RB datasets have a stronger personalization, in which multi-view ratings are more valuable than side information for recommendations.

\section{Conclusion}
DTTF is proposed for cross-domain multi-view recommendation
by combining tensor factorization and deep structure
in both source and target domains. Private latent factors link
with deep structures while view latent factor is taken a bridge
between domains, which are learned by jointly optimizing
tensor factorization and SDAEs. Experimental results on the
real-world datasets show that our proposed approach achieves
a superiority compared with state-of-the-art works.

\bibliography{ref}

\begin{thebibliography}{10}
\providecommand{\url}[1]{#1}
\csname url@samestyle\endcsname
\providecommand{\newblock}{\relax}
\providecommand{\bibinfo}[2]{#2}
\providecommand{\BIBentrySTDinterwordspacing}{\spaceskip=0pt\relax}
\providecommand{\BIBentryALTinterwordstretchfactor}{4}
\providecommand{\BIBentryALTinterwordspacing}{\spaceskip=\fontdimen2\font plus
\BIBentryALTinterwordstretchfactor\fontdimen3\font minus
  \fontdimen4\font\relax}
\providecommand{\BIBforeignlanguage}[2]{{%
\expandafter\ifx\csname l@#1\endcsname\relax
\typeout{** WARNING: IEEEtran.bst: No hyphenation pattern has been}%
\typeout{** loaded for the language `#1'. Using the pattern for}%
\typeout{** the default language instead.}%
\else
\language=\csname l@#1\endcsname
\fi
#2}}
\providecommand{\BIBdecl}{\relax}
\BIBdecl

\bibitem{1}
Y.~Koren, R.~Bell, and C.~Volinsky, ``Matrix factorization techniques for
  recommender systems,'' \emph{Computer}, vol.~8, pp. 30--37, 2009.

\bibitem{2}
E.~E. Papalexakis, C.~Faloutsos, and N.~D. Sidiropoulos, ``Tensors for data
  mining and data fusion: Models, applications, and scalable algorithms,''
  \emph{ACM Transactions on Intelligent Systems and Technology (TIST)}, vol.~8,
  no.~2, pp. 1--44, 2016.

\bibitem{3}
K.~Lakiotaki, N.~F. Matsatsinis, and A.~Tsoukias, ``Multicriteria user modeling
  in recommender systems,'' \emph{IEEE Intelligent Systems}, vol.~26, no.~2,
  pp. 64--76, 2011.

\bibitem{4}
K.~Lakiotaki, S.~Tsafarakis, and N.~Matsatsinis, ``Uta-rec: a recommender
  system based on multiple criteria analysis,'' in \emph{Proceedings of the
  2008 ACM conference on Recommender systems}, 2008, pp. 219--226.

\bibitem{5}
N.~Sahoo, R.~Krishnan, G.~Duncan, and J.~Callan, ``Research note—the halo
  effect in multicomponent ratings and its implications for recommender
  systems: The case of yahoo! movies,'' \emph{Information Systems Research},
  vol.~23, no.~1, pp. 231--246, 2011.

\bibitem{6}
A.~Mikeli, D.~Apostolou, and D.~Despotis, ``A multi-criteria recommendation
  method for interval scaled ratings,'' in \emph{2013 IEEE/WIC/ACM
  International Joint Conferences on Web Intelligence (WI) and Intelligent
  Agent Technologies (IAT)}, vol.~3.\hskip 1em plus 0.5em minus 0.4em\relax
  IEEE, 2013, pp. 9--12.

\bibitem{7}
A.~Karatzoglou, X.~Amatriain, L.~Baltrunas, and N.~Oliver, ``Multiverse
  recommendation: n-dimensional tensor factorization for context-aware
  collaborative filtering,'' in \emph{Proceedings of the fourth ACM conference
  on Recommender systems}, 2010, p.~79.

\bibitem{8}
Z.~Chen and D.~Wang, ``Multi-initialization meta-learning with domain
  adaptation,'' in \emph{ICASSP 2021-2021 IEEE International Conference on
  Acoustics, Speech and Signal Processing (ICASSP)}.\hskip 1em plus 0.5em minus
  0.4em\relax IEEE, 2021, pp. 1390--1394.

\bibitem{9}
P.~Bhargava, T.~Phan, J.~Zhou, and J.~Lee, ``Who, what, when, and where:
  Multi-dimensional collaborative recommendations using tensor factorization on
  sparse user-generated data,'' in \emph{Proceedings of the 24th international
  conference on world wide web}, 2015, pp. 130--140.

\bibitem{10}
L.~Yao, Q.~Z. Sheng, Y.~Qin, X.~Wang, A.~Shemshadi, and Q.~He, ``Context-aware
  point-of-interest recommendation using tensor factorization with social
  regularization,'' in \emph{Proceedings of the 38th international ACM SIGIR
  conference on research and development in information retrieval}, 2015, pp.
  1007--1010.

\bibitem{11}
X.~Dong, L.~Yu, Z.~Wu, Y.~Sun, L.~Yuan, and F.~Zhang, ``A hybrid collaborative
  filtering model with deep structure for recommender systems,'' in
  \emph{Proceedings of the AAAI Conference on artificial intelligence}, 2017,
  pp. 1309--1315.

\bibitem{12}
Z.~Chen, T.~Xiao, and K.~Kuang, ``Ba-gnn: On learning bias-aware graph neural
  network,'' in \emph{2022 IEEE 38th International Conference on Data
  Engineering (ICDE)}.\hskip 1em plus 0.5em minus 0.4em\relax IEEE, 2022, pp.
  3012--3024.

\bibitem{13}
H.~Wang, N.~Wang, and D.-Y. Yeung, ``Collaborative deep learning for
  recommender systems,'' in \emph{Proceedings of the 21th ACM SIGKDD
  international conference on knowledge discovery and data mining}, 2015, pp.
  1235--1244.

\bibitem{14}
T.~Xiao, Z.~Chen, D.~Wang, and S.~Wang, ``Learning how to propagate messages in
  graph neural networks,'' in \emph{Proceedings of the 27th ACM SIGKDD
  Conference on Knowledge Discovery \& Data Mining}, 2021, pp. 1894--1903.

\bibitem{15}
P.~Zhu, Q.~Hu, Q.~Hu, C.~Zhang, and Z.~Feng, ``Multi-view label embedding,''
  \emph{Pattern recognition}, vol.~84, pp. 126--135, 2018.

\bibitem{16}
S.~Huang, Z.~Kang, I.~W. Tsang, and Z.~Xu, ``Auto-weighted multi-view
  clustering via kernelized graph learning,'' \emph{Pattern Recognition},
  vol.~88, pp. 174--184, 2019.

\bibitem{17}
Y.~Zhang, Y.~Yang, T.~Li, and H.~Fujita, ``A multitask multiview clustering
  algorithm in heterogeneous situations based on lle and le,''
  \emph{Knowledge-Based Systems}, vol. 163, pp. 776--786, 2019.

\bibitem{18}
Z.~Chen, J.~Ge, H.~Zhan, S.~Huang, and D.~Wang, ``Pareto self-supervised
  training for few-shot learning,'' in \emph{Proceedings of the IEEE/CVF
  Conference on Computer Vision and Pattern Recognition}, 2021, pp.
  13\,663--13\,672.

\bibitem{19}
H.~Cao, S.~Bernard, R.~Sabourin, and L.~Heutte, ``Random forest dissimilarity
  based multi-view learning for radiomics application,'' \emph{Pattern
  Recognition}, vol.~88, pp. 185--197, 2019.

\bibitem{20}
Y.~Yuan, G.~Xun, K.~Jia, and A.~Zhang, ``A multi-view deep learning framework
  for eeg seizure detection,'' \emph{IEEE journal of biomedical and health
  informatics}, vol.~23, no.~1, pp. 83--94, 2018.

\bibitem{21}
S.~Huang, Z.~Kang, and Z.~Xu, ``Auto-weighted multi-view clustering via deep
  matrix decomposition,'' \emph{Pattern Recognition}, vol.~97, p. 107015, 2020.

\bibitem{22}
N.~Zhang, S.~Ding, T.~Sun, H.~Liao, L.~Wang, and Z.~Shi, ``Multi-view rbm with
  posterior consistency and domain adaptation,'' \emph{Information Sciences},
  vol. 516, pp. 142--157, 2020.

\bibitem{23}
A.~P. Singh and G.~J. Gordon, ``Relational learning via collective matrix
  factorization,'' in \emph{Proceedings of the 14th ACM SIGKDD international
  conference on Knowledge discovery and data mining}, 2008, pp. 650--658.

\bibitem{24}
S.~Gai, F.~Zhao, Y.~Kang, Z.~Chen, D.~Wang, and A.~Tang, ``Deep transfer
  collaborative filtering for recommender systems,'' in \emph{PRICAI 2019:
  Trends in Artificial Intelligence: 16th Pacific Rim International Conference
  on Artificial Intelligence, Cuvu, Yanuca Island, Fiji, August 26-30, 2019,
  Proceedings, Part III 16}.\hskip 1em plus 0.5em minus 0.4em\relax Springer,
  2019, pp. 515--528.

\bibitem{25}
G.~Hu, Y.~Zhang, and Q.~Yang, ``Mtnet: a neural approach for cross-domain
  recommendation with unstructured text,'' \emph{KDD Deep Learning Day}, pp.
  1--10, 2018.

\bibitem{26}
G.~Adomavicius and Y.~Kwon, ``New recommendation techniques for multicriteria
  rating systems,'' \emph{IEEE Intelligent Systems}, vol.~22, no.~3, pp.
  48--55, 2007.

\bibitem{27}
Z.~Chen, D.~Wang, and S.~Yin, ``Improving cold-start recommendation via
  multi-prior meta-learning,'' in \emph{Advances in Information Retrieval: 43rd
  European Conference on IR Research, ECIR 2021, Virtual Event, March 28--April
  1, 2021, Proceedings, Part II 43}.\hskip 1em plus 0.5em minus 0.4em\relax
  Springer, 2021, pp. 249--256.

\bibitem{28}
D.~Jannach, Z.~Karakaya, and F.~Gedikli, ``Accuracy improvements for
  multi-criteria recommender systems,'' in \emph{Proceedings of the 13th ACM
  conference on electronic commerce}, 2012, pp. 674--689.

\bibitem{29}
Y.~Zheng, ``Criteria chains: a novel multi-criteria recommendation approach,''
  in \emph{Proceedings of the 22nd International Conference on Intelligent User
  Interfaces}, 2017, pp. 29--33.

\bibitem{30}
M.~Hassan and M.~Hamada, ``Genetic algorithm approaches for improving
  prediction accuracy of multi-criteria recommender systems,''
  \emph{International Journal of Computational Intelligence Systems}, vol.~11,
  no.~1, pp. 146--162, 2018.

\bibitem{31}
A.~M. Turk and A.~Bilge, ``Robustness analysis of multi-criteria collaborative
  filtering algorithms against shilling attacks,'' \emph{Expert Systems with
  Applications}, vol. 115, pp. 386--402, 2019.

\bibitem{32}
S.~Wang, J.~Yang, Z.~Chen, H.~Yuan, J.~Geng, and Z.~Hai, ``Global and local
  tensor factorization for multi-criteria recommender system,''
  \emph{Patterns}, vol.~1, no.~2, p. 100023, 2020.

\bibitem{33}
Y.~Zheng, S.~Shekhar, A.~A. Jose, and S.~K. Rai, ``Integrating
  context-awareness and multi-criteria decision making in educational
  learning,'' in \emph{Proceedings of the 34th ACM/SIGAPP Symposium on Applied
  Computing}, 2019, pp. 2453--2460.

\bibitem{34}
P.~Li and A.~Tuzhilin, ``Latent multi-criteria ratings for recommendations,''
  in \emph{Proceedings of the 13th ACM Conference on Recommender Systems},
  2019, pp. 428--431.

\bibitem{35}
N.~Nassar, A.~Jafar, and Y.~Rahhal, ``A novel deep multi-criteria collaborative
  filtering model for recommendation system,'' \emph{Knowledge-Based Systems},
  vol. 187, p. 104811, 2020.

\bibitem{36}
Z.~Su, J.~Yan, H.~Ling, and H.~Chen, ``Research on personalized recommendation
  algorithm based on ontological user interest model,'' \emph{Journal of
  Computational Information Systems}, vol.~8, no.~1, pp. 169--181, 2012.

\bibitem{37}
M.~Nilashi, O.~bin Ibrahim, and N.~Ithnin, ``Hybrid recommendation approaches
  for multi-criteria collaborative filtering,'' \emph{Expert Systems with
  Applications}, vol.~41, no.~8, pp. 3879--3900, 2014.

\bibitem{38}
Q.~Li, C.~Wang, and G.~Geng, ``Improving personalized services in mobile
  commerce by a novel multicriteria rating approach,'' in \emph{Proceedings of
  the 17th international conference on World Wide Web}, 2008, pp. 1235--1236.

\bibitem{39}
S.~Rendle, L.~Balby~Marinho, A.~Nanopoulos, and L.~Schmidt-Thieme, ``Learning
  optimal ranking with tensor factorization for tag recommendation,'' in
  \emph{Proceedings of the 15th ACM SIGKDD international conference on
  Knowledge discovery and data mining}, 2009, pp. 727--736.

\bibitem{40}
S.~Rendle, Z.~Gantner, C.~Freudenthaler, and L.~Schmidt-Thieme, ``Fast
  context-aware recommendations with factorization machines,'' in
  \emph{Proceedings of the 34th international ACM SIGIR conference on Research
  and development in Information Retrieval}, 2011, pp. 635--644.

\bibitem{41}
Z.~Zhang and S.~Aeron, ``Exact tensor completion using t-svd,'' \emph{IEEE
  Transactions on Signal Processing}, vol.~65, no.~6, pp. 1511--1526, 2017.

\bibitem{42}
Z.~Chen, S.~Gai, and D.~Wang, ``Deep tensor factorization for multi-criteria
  recommender systems,'' in \emph{2019 IEEE International Conference on Big
  Data (Big Data)}.\hskip 1em plus 0.5em minus 0.4em\relax IEEE, 2019, pp.
  1046--1051.

\bibitem{43}
D.~Jannach, M.~Zanker, and M.~Fuchs, ``Leveraging multi-criteria customer
  feedback for satisfaction analysis and improved recommendations,''
  \emph{Information Technology \& Tourism}, vol.~14, no.~2, pp. 119--149, 2014.

\bibitem{44}
S.~Li, J.~Kawale, and Y.~Fu, ``Deep collaborative filtering via marginalized
  denoising auto-encoder,'' in \emph{Proceedings of the 24th ACM international
  on conference on information and knowledge management}, 2015, pp. 811--820.

\bibitem{45}
J.~McAuley, J.~Leskovec, and D.~Jurafsky, ``Learning attitudes and attributes
  from multi-aspect reviews,'' in \emph{2012 IEEE 12th International Conference
  on Data Mining}.\hskip 1em plus 0.5em minus 0.4em\relax IEEE, 2012, pp.
  1020--1025.

\bibitem{46}
X.~He, T.~Chen, M.-Y. Kan, and X.~Chen, ``Trirank: Review-aware explainable
  recommendation by modeling aspects,'' in \emph{Proceedings of the 24th ACM
  international on conference on information and knowledge management}, 2015,
  pp. 1661--1670.

\end{thebibliography}
\bibliographystyle{IEEEtran}


\end{document}